\documentclass{article}

\usepackage{microtype}
\usepackage{graphicx}
\usepackage{booktabs} 
\RequirePackage{xcolor}

\usepackage{hyperref}

\usepackage[accepted]{icml2021}

\usepackage[utf8]{inputenc} 
\usepackage[T1]{fontenc}    
\usepackage{hyperref}       
\usepackage{url}            
\usepackage{booktabs}       
\usepackage{amsfonts}       
\usepackage{nicefrac}       
\usepackage{microtype}      

\usepackage{times}
\usepackage{latexsym}
\usepackage{enumitem}

\usepackage{graphicx}
\usepackage{amssymb}
\usepackage{comment}

\usepackage{caption}
\usepackage{subcaption}
\usepackage{booktabs}

\usepackage{multicol}
\usepackage{multirow}
\usepackage{array, tabularx,  ragged2e,  booktabs}
\usepackage{natbib}

\usepackage[utf8]{inputenc}
\usepackage{graphicx}
\usepackage{pgfplots}
\pgfplotsset{compat=1.14}

\usepackage{ctable}
\usepackage{wrapfig}
\usepackage{tikz}
\usetikzlibrary{positioning,arrows,trees,shapes,fit,shadows}

\usepackage{amsfonts}
\usepackage{ifthen}
\usepackage{booktabs}
\usepackage{amssymb}
\usepackage{supertabular}
\usepackage{array}
\usepackage{multirow}
\usepackage{fancyhdr}
\usepackage[normalem]{ulem}

\usepackage{amssymb}
\usepackage{supertabular}
\usepackage{array}
\usepackage{multirow}
\usepackage{fancyhdr}
\usepackage{psfrag}
\usepackage{amsmath}
\usepackage{hhline}
\usepackage{type1cm}
\usepackage{lettrine}
\usepackage[colorinlistoftodos,prependcaption,textsize=tiny]{todonotes}

\usepackage{algorithm}
\usepackage{algorithmic} 

\usepackage{xspace}

\usepackage{babel}
\usepackage[export]{adjustbox}
\usepackage{cleveref}
\usepackage{scrextend}

\crefformat{section}{\S#2#1#3} 
\crefformat{subsection}{\S#2#1#3}
\crefformat{subsubsection}{\S#2#1#3}

\usepackage{amsmath,amsfonts,bm}

\def\eqref#1{equation~\ref{#1}}

\def\1{\bm{1}}

\makeatletter
\newcommand{\xRightarrow}[2][]{\ext@arrow 0359\Rightarrowfill@{#1}{#2}}
\makeatother

\DeclareMathAlphabet{\mathsfit}{\encodingdefault}{\sfdefault}{m}{sl}
\SetMathAlphabet{\mathsfit}{bold}{\encodingdefault}{\sfdefault}{bx}{n}

\def\gJ{{\mathcal{J}}}

\def\sX{{\mathbb{X}}}

\DeclareMathOperator*{\argmax}{arg\,max}

\newcommand{\ie}{\emph{i.e.,}\xspace}

\newcommand{\Ni}{({\em i})~}
\newcommand{\Nii}{({\em ii})~}
\newcommand{\Niii}{({\em iii})~}

\usepackage{float}

\DeclareCaptionLabelFormat{andtable}{#1~#2  \&  \tablename~\thetable}

\usepackage{tikz-cd}
\usepackage{mathtools}
\usepackage{todonotes}

\icmltitlerunning{Cross-model Back-translated Distillation for Unsupervised Machine Translation}

\begin{document}

\twocolumn[
\icmltitle{Cross-model Back-translated Distillation\\for Unsupervised Machine Translation}




\icmlsetsymbol{equal}{*}

\begin{icmlauthorlist}
\icmlauthor{Xuan-Phi Nguyen}{to,goo}
\icmlauthor{Shafiq Joty}{to,ed}
\icmlauthor{Thanh-Tung Nguyen}{to,goo}
\icmlauthor{Wu Kui}{goo}
\icmlauthor{Ai Ti Aw}{goo}
\end{icmlauthorlist}

\icmlaffiliation{to}{Nanyang Technological University}
\icmlaffiliation{goo}{Institute for Infocomm Research (I$^2$R), A*STAR}
\icmlaffiliation{ed}{Salesforce Research Asia}

\icmlcorrespondingauthor{Xuan-Phi Nguyen}{nguyenxu002@e.ntu.edu.sg}

\icmlkeywords{Machine Learning, ICML}

\vskip 0.3in
]




\printAffiliationsAndNotice{}  

\begin{abstract}
Recent unsupervised machine translation (UMT) systems usually employ three main principles: initialization, language modeling and iterative back-translation, though they may apply them differently. Crucially, iterative back-translation and denoising auto-encoding for language modeling provide data diversity to train the UMT systems. However, the gains from these diversification processes has seemed to plateau. We introduce a novel component to the standard UMT framework called Cross-model Back-translated Distillation (CBD), that is aimed to induce another level of data diversification that existing principles lack. CBD is applicable to all previous UMT approaches. In our experiments, CBD achieves the state of the art in the {WMT'14 English-French}, WMT'16 English-German and English-Romanian bilingual unsupervised translation tasks, with 38.2, 30.1, and 36.3 BLEU respectively. It also yields 1.5--3.3 BLEU improvements in IWSLT English-French and English-German tasks. Through extensive experimental analyses, we show that CBD is effective because it embraces data diversity while other similar variants do not.
\end{abstract}

\section{Introduction}\label{sec:intro}

Machine translation (MT) is a core task in natural language processing  that involves both language understanding and generation. Recent neural approaches \citep{vaswani2017attention,payless_wu2018} have advanced the state of the art with near human-level performance \citep{hassanawadalla2018achieving}. 
However, they continue to rely heavily on large parallel data.
As a result, the search for unsupervised alternatives using only monolingual data has been active. While \citet{ravi2011deciphering} and \citet{smt_without_parallel_klementiev2012toward} proposed various unsupervised techniques for statistical MT (SMT), \citet{lample2017unsupervised,lample2018phrase_unsup} established a general framework for modern unsupervised MT (UMT) that works for both SMT and neural MT (NMT) models. The framework has three main principles: model initialization, language modeling and iterative back-translation.
Model initialization bootstraps the model with a knowledge prior like word-level transfer \citep{muse_conneau2017word}. Language modeling, which takes the form of denoising auto-encoding (DAE) in NMT \citep{lample2018phrase_unsup}, trains the model to generate plausible sentences in a language. Meanwhile, iterative back-translation (IBT) facilitates cross-lingual translation training by generating noisy source sentences for original target sentences.
The {recent approaches}  differ on how they apply each of these three principles. For instance, \citet{lample2017unsupervised} use an unsupervised word-translation model \citep{muse_conneau2017word} for model initialization, while \citet{conneau2019cross_xlm} use a pretrained cross-lingual masked language model (XLM).

{In this paper, we focus on a different aspect of the UMT framework, namely, its \emph{data diversification} process. In this context, we refer data diversification as only sentence level variations, and not contextual topics or genres. If we look from this view, the DAE and IBT steps of the UMT framework also perform some form of data diversification to train the model.} 
Specifically, the noise model in the DAE process generates \emph{new}, but noised, versions of the input data, which are used to train the model with a reconstruction objective. Likewise, the IBT step involves the same
UMT model to create synthetic parallel pairs (with the source being synthetic), which are then used to train the model. Since the NMT model is updated with DAE and IBT simultaneously, the model generates \emph{fresh} translations in each back-translation step. Overall, thanks to DAE and IBT, the model gets better at translating by iteratively training on the newly created and diversified data {whose quality also improves over time.} This argument also applies to statistical UMT, except for the lack of the DAE \citep{lample2018phrase_unsup}. However, we conjecture that these diversification methods may have reached their limit as the performance does not improve further the longer we train the UMT models.

In this work, we introduce a fourth principle to the standard UMT framework: Cross-model Back-translated Distillation\footnote{{Code: {\href{https://github.com/nxphi47/multiagent_crosstranslate}{https://github.com/nxphi47/multiagent\_crosstranslate}.}}} or CBD (\cref{sec:method}), with the aim to induce another level of diversification that the existing UMT principles lack. CBD initially trains two bidirectional UMT agents (models) using existing approaches. Then, one of the two agents translates the monolingual data from one language $s$ to another $t$ in the first level. In the second level, the generated data are back-translated from $t$ to $s$ by the \emph{other agent}. In the final step, the synthetic parallel data created by the first and second levels are used to distill a supervised MT model. CBD is applicable to any existing UMT method {and is more efficient than ensembling approaches \citep{ensemble_distill_freitag2017}} (\cref{subsec:alternatives1}).

In the experiments (\cref{sec:experiments}), CBD establishes the state of the art (SOTA) in the bilingual unsupervised translation tasks of WMT'14 English-French, WMT'16 English-German and WMT'16 English-Romanian, with {38.2, 30.1 and 36.3 BLEU, respectively}.
Without large scale pretrained models and data, our method shows consistent improvements of 1.0-2.0 BLEU compared to the baselines in these tasks. It also boosts the performance on IWSLT'14 English-German and IWSLT'13 English-French tasks significantly. {In our analysis, we explain with experiments why other similar variants (\cref{subsec:cross_translation}) {and other alternatives from the literature (\cref{subsec:alternatives2})} do not work well and cross-model back-translation is crucial for our method.} We further demonstrate that CBD enhances the baselines by achieving greater diversity as measured by back-translation BLEU (\cref{subsec:data_diversity}). 
\section{Background}\label{sec:background}

\citet{ravi2011deciphering} were among the first to propose a UMT system by framing the problem as a \textit{decipherment} task that considers non-English text as a cipher for English. Nonetheless, the method is limited and may not be applicable to the current well-established NMT systems \citep{luong2015effective,vaswani2017attention,payless_wu2018}. \citet{lample2017unsupervised} set the foundation for modern UMT. They propose to maintain two encoder-decoder networks simultaneously for both source and target languages, and train them via denoising auto-encoding, iterative back-translation and adversarial training. 
{In their follow-up work,  \citet{lample2018phrase_unsup} formulate a common UMT framework for both Phrase-based SMT (PBSMT) and NMT with three basic principles that can be customized.} Specifically, the three main principles of UMT are:   


\vspace{-1em}
\begin{itemize}[itemsep=-0.3pt,leftmargin=*]
    \item \textbf{Initialization}: A non-randomized {cross- or multi-lingual} initialization that represents a knowledge prior to bootstrap the UMT model. For instance, \citet{lample2017unsupervised} and \citet{effective-artetxe-etal-2019} use an {unsupervised} word-translation model MUSE \citep{muse_conneau2017word} as initialization to promote word-to-word {cross-lingual} transfer. {\cite{lample2018phrase_unsup} use a shared jointly trained sub-word \citep{sennrich2015neural} dictionary.} On the other hand, \citet{conneau2019cross_xlm} use a pretrained cross-lingual masked language model (XLM) to initialize the unsupervised NMT model. 
    
    \item \textbf{Language modeling}: Training a language model on monolingual data helps the UMT model to generate fluent texts. The neural UMT approaches \citep{lample2017unsupervised,lample2018phrase_unsup,conneau2019cross_xlm} use denoising auto-encoder training to achieve language modeling effects in the neural model. Meanwhile, the PBSMT variant proposed by \citet{lample2018phrase_unsup} uses the KenLM smoothed n-gram language models \citep{pbsmt_lm_heafield2011kenlm}.

    \item \textbf{Iterative back-translation}: Back-translation \citep{backtranslate_sennrich-etal-2016-improving} brings about the bridge between source and target languages by using a backward model that translates data from target to source. The {(source and target)} monolingual data is translated back and forth {iteratively} to {progress} the UMT model in both directions.
\end{itemize}
\vspace{-1em}

During training, the initialization step is conducted once, while the denoising and back-translation steps are often executed in an alternating manner.\footnote{{The KenLM language model in PBSMT \citep{lample2018phrase_unsup} was kept fixed during the training process.}} It is worth noting that depending on different implementations, the parameters for backward and forward components may be separate \citep{lample2017unsupervised} or shared \citep{lample2018phrase_unsup,conneau2019cross_xlm}. A {parameter-shared} cross-lingual NMT model has the capability to translate from either source or target, while a UMT system with parameter-separate models has to maintain two models. 
Either way, we deem a standard UMT system to be bidirectional, i.e., it is capable of translating from either source or target language.


Our proposed cross-model back-translated distillation (CBD) works outside this well-established framework. It employs two UMT agents to create extra diversified data apart from what existing methods already offer, rendering it a useful add-on to the general UMT framework. Furthermore, different implementations of UMT as discussed above can be plugged into the CBD system to achieve a performance boost, even for future methods that may potentially not employ the three principles.

\section{Cross-model Back-translated Distillation} \label{sec:method}

\begin{figure}
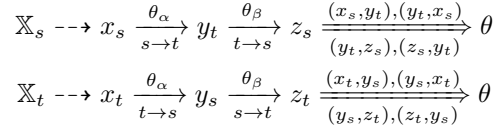

\begin{center}
\begin{equation*}
    \sX_s \dashrightarrow x_s  \xrightarrow[s \rightarrow t]{\theta_{\alpha}}  y_t \xrightarrow[t \rightarrow s]{\theta_{\beta}} z_s \xRightarrow[(y_t,z_s),(z_s,y_t)]{(x_s,y_t),(y_t,x_s)} \theta
\end{equation*}
\vspace{-1em}
\begin{equation*}
    \sX_t \dashrightarrow x_t  \xrightarrow[t \rightarrow s]{\theta_{\alpha}}  y_s \xrightarrow[s \rightarrow t]{\theta_{\beta}} z_t \xRightarrow[(y_s,z_t),(z_t,y_s)]{(x_t,y_s),(y_s,x_t)} \theta
\end{equation*}
\vspace{-1em}
\caption{The sampling process of $x_s,y_t,z_s,x_t,y_s,z_t$. The variable ordered set $(\theta_{\alpha},\theta_{\beta})$ is replaced with $(\theta_1,\theta_2)$ and $(\theta_2,\theta_1)$ iteratively in \Cref{alg:macd}. All synthetic parallel pairs are used to train $\theta$ in a supervised way.}
\label{fig:cbd:translation_diagram} 
\end{center}
\end{figure}

In section, we explain our CBD method in more details. Specifically, let $\sX_{s}$ and $\sX_{t}$ denote the two sets of monolingual data for languages $s$ and $t$, respectively. We first train two UMT agents independently with two different parameter sets $\theta_1$ and $\theta_2$ using existing methods \citep{lample2017unsupervised,lample2018phrase_unsup,conneau2019cross_xlm}.\footnote{For neural approaches, changing the random seeds would do the trick, {while PBSMT methods would need to randomize the initial embeddings and/or subsample the training data.}} 
Since a UMT agent with parameter set ${\theta_i} \in \{\theta_1, \theta_2\}$ is deemed \textit{bidirectional} in our setup, we denote $y_t \sim P(\cdot|x_s, {\theta_i})$ to be a translation sample from language $s$ to $t$ of input sentence $x_s$ using model ${\theta_i}$.
Assuming $\Theta = \{ \theta_1, \theta_2 \}$, we then define $x_s \sim \sX_s$, $y_t \sim P(\cdot|x_s, {\theta_\alpha})$ and $z_s \sim P(\cdot|y_t, {\theta_\beta})$ to be a sample $x_s$ from $\sX_{s}$, a translation of $x_s$ to language $t$ using model $\theta_\alpha$, and a translation of $y_t$ back to language $s$ using $\theta_\beta$, respectively, {with $\theta_{\alpha}$ being either $\theta_1$ or $\theta_2$ and $\theta_{\beta}=\Theta \setminus \theta_{\alpha}$.} Note that in this formulation, the model $\theta_\alpha$ that produces $y_t$ is different from the one $\theta_\beta$ that produces $z_s$. Similarly, we define $x_t \sim \sX_t$, $y_s \sim P(\cdot|x_t, {\theta_\alpha})$ and $z_t \sim P(\cdot|y_s,\theta_\beta)$ in the same manner for $\sX_t$. \Cref{fig:cbd:translation_diagram} further illustrates this process.  


With these generated samples, we train a \emph{supervised} MT model parameterized by $\theta$ to maximize the joint probabilities of the aforementioned six random variables, \ie\ $x_s,y_t,z_s,x_t,y_s$ and $z_t$. Equivalently, we minimize the following derived objective function:
\begin{equation}
\begin{split}
\hspace{-0.5em}& \gJ(\theta) = \frac{1}{2} \Big[ -\log P_{\theta}(y_t|z_s) - \log P_{\theta}(y_t|x_s) \\
\hspace{-0.5em}& - \log P_{\theta}(z_s|y_t) - \log P_{\theta}(x_s|y_t) - \log P_{\theta}(y_s|z_t)  \\
\hspace{-0.5em}& - \log P_{\theta}(y_s|x_t) - \log P_{\theta}(z_t|y_s) - \log P_{\theta}(y_s|x_t)  \Big]
\end{split}
\hspace{-0.5em}
\label{eqn:nll:out}
\end{equation}
Mathematical derivations and detailed explanations of objective $\gJ(\theta)$ are further given in the Appendix. Considering the sampling process of $x_s,y_s,z_s,x_t,y_t$ and $z_t$, The model $\theta$ is trained by minimizing the following CBD loss function:
\begin{equation}
\begin{split}
\hspace{-0.3em}\mathcal{L_{\theta}}(\theta_{\alpha},\theta_{\beta}) =\hspace{-0.7em} \displaystyle \mathop{\mathbb{E}}_{\substack{z_s \sim P(\cdot|y_t,\theta_{\beta}), y_t \sim P(\cdot|x_s,\theta_{\alpha}), x_s \sim \sX_s\\z_t \sim P(\cdot|y_s,\theta_{\beta}), y_s \sim P(\cdot|x_t, \theta_{\alpha}), x_t \sim \sX_t}} [ \gJ (\theta) ] \hspace{-0.7em} \label{eqn:cbd_loss}
\end{split}
\end{equation}
where $\theta_{\alpha},\theta_{\beta} \in \Theta$ are the given UMT models; $\theta_{\alpha}$ is used to generate $y_t$ and $y_s$ from $x_s$ and $x_t$ respectively, while $\theta_{\beta}$ is used to back-translate $y_t$ and $y_s$ to $z_s$ and $z_t$ respectively. \Cref{alg:macd} describes the overall CBD training process, where the ordered pair $(\theta_{\alpha},\theta_{\beta})$ is alternated between $(\theta_1,\theta_2)$ and $(\theta_2,\theta_1)$

\begin{algorithm}[t!]
\caption{Cross-model Back-translated Distillation: Given monolingual data $\sX_{s}$ and $\sX_{t}$ of languages $s$ and $t$, return a UMT model with parameters $\theta$.}
\label{alg:macd}
\begin{algorithmic}[1]
    \STATE Train the 1st UMT agent with parameters $\theta_1$
    \STATE Train the 2nd UMT agent with parameters $\theta_2$
    \STATE Initialize model $\theta$ (randomly or with pretrained model)
    \WHILE{until convergence}
        \STATE $\theta \gets \theta - \eta \nabla_{\theta} \mathcal{L_\theta}(\theta_{\alpha}=\theta_1,\theta_{\beta}=\theta_2)$
        \STATE $\theta \gets \theta - \eta \nabla_{\theta} \mathcal{L_\theta}(\theta_{\alpha}=\theta_2,\theta_{\beta}=\theta_1)$
    \ENDWHILE
    \STATE \textbf{return} $\theta$
\end{algorithmic}
\end{algorithm}


To describe the CBD strategy more conceptually, in each iteration step of \Cref{alg:macd}, each agent ${\theta_\alpha} \in \{\theta_1,\theta_2\}$ generates {translations} from the monolingual data $\sX_s$ and $\sX_t$ of both languages $s$ and $t$ to acquire the \emph{first level} of synthetic parallel data $(x_s, y_t)$ and $(x_t, y_s)$. In the \emph{second level}, the other agent $\theta_\beta = \{\theta_1,\theta_2\} \setminus \theta_\alpha$ is used to generate the translation $z_s$ of the translation $y_t$ of $x_s$ (and similarly for $z_t$ from the translation $y_s$ of $x_t$). This process is basically \emph{back-translation}, but with the backward model coming from a different regime than that of the forward model. The fact that the first level agent must be different from the second level agent is crucial to achieve the desirable level of diversity in data generation. After this, we update the model $\theta$ using {all the} synthetic pairs $(x,y)$ and $(y,z)$ using the objective function defined in \Cref{eqn:nll:out}.

In this way, firstly, the model $\theta$ gets trained on the translated products $\{ (x_s \leftrightarrow y_t), (x_t \leftrightarrow y_s)\}$ of the UMT teachers, making it as capable as the teachers. Secondly, the model $\theta$ is also trained on the second-level data $\{ (y_t \leftrightarrow z_s), (y_s \leftrightarrow z_t)\}$ which is slightly different from the first-level data. Thus, this mechanism provides extra data diversification to the system $\theta$ in addition to what the UMT teachers already offer, resulting in our final model outperforming the UMT baselines (\cref{sec:experiments}). However, one may argue that since $\theta_1$ and $\theta_2$ are trained in a similar fashion, $z$ will be the same as $x$, resulting in a duplicate pair. In our experiments, on the contrary, the back-translated dataset contains only around 14\% duplicates across different language pairs, as shown in our analysis on data diversity in \cref{subsec:data_diversity}. 

In the Appendix, we provide a more generalized version of CBD with $n$ ($\ge2$) UMT agents, where we also analyze its effectiveness on the IWSLT translation tasks.

\section{Experiments}\label{sec:experiments}


We present our experiments on the large scale WMT (\cref{subsec:wmt_large}) and base WMT (\cref{subsec:wmt}) tasks, followed by IWSLT (\cref{subsec:iwslt}).


\vspace{-0.5em}

\subsection{Large scale WMT experiments}\label{subsec:wmt_large}

\paragraph{Setup.} We use the codebase from \citet{conneau2019cross_xlm} and follow exactly their model setup. Specifically, we use all of the monolingual data from 2007-2017 WMT News Crawl datasets, which yield 190M, 78M, 309M and 3M sentences for language English (En), French (Fr), German (De) and Romanian (Ro) respectively. We filter out sentences whose lengths are over 175 tokens. For each language pair, we build a jointly bilingual dictionary of 60K sub-word units using Byte-Pair Encoding \citep{sennrich2015neural}. To save resource, we reuse the pretrained XLM \cite{conneau2019cross_xlm} and MASS\footnote{MASS outperforms XLM in our Romanian-related experiments.} \cite{song2019mass} UMT finetuned models as our $\theta_1$ and $\theta_2$ respectively. We initialize the CBD supervised MT model $\theta$ with the pretrained XLM model provided by \citet{conneau2019cross_xlm} for En-Fr and De-En pairs and the pretrained MASS model from \citet{song2019mass} for En-Ro pairs, both of which are Transformers with 6 layers and 1024 dimensions. 
We train the model with a 2K tokens per batch on a 8-GPU system. Like all previous work, we evaluate the models using the tokenized Moses \href{https://github.com/moses-smt/mosesdecoder/blob/master/scripts/generic/multi-bleu.perl}{\emph{multi-bleu.perl} script} \citep{koehn2007moses}.

\begin{figure}[t!]
    \centering
    \includegraphics[width=\columnwidth]{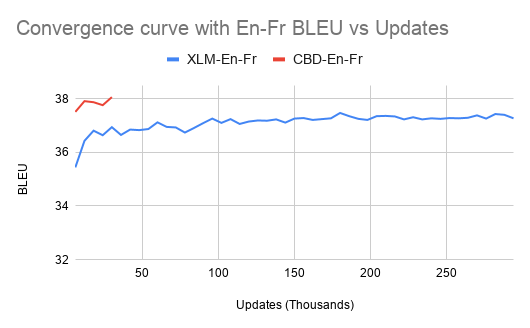}
    \caption{Convergence speed of CBD in comparison with baseline XLM, represented by BLEU score on the WMT'14 En-Fr testset after a number of training updates. Analyses of other languages are given in the Appendix.}
    \label{fig:convergence}
    \vspace{-1em}
\end{figure}

\vspace{-1em}
\paragraph{Results.} Table \ref{table:large_results} shows the performance of CBD in comparison with recent UMT methods. Our method establishes the SOTA in the WMT unsupervised tasks, with 38.2, 35.5, 30.1, 36.3, {36.3 and 33.8} for the large scale WMT En-Fr, Fr-En, En-De, De-En, En-Ro and Ro-En tasks respectively. This translates to up to 1.8 BLEU improvements over the previous SOTA  \citep{song2019mass}. More interestingly, given that the hard work in training the teacher and initial models $\theta_1$, $\theta_2$ and $\theta$ has been done by \citet{conneau2019cross_xlm} and \citet{song2019mass}, our CBD requires a fraction of additional resource to outperform the baselines. This is illustrated in \Cref{fig:convergence}, where CBD only needs around 20K updates to converge while the baseline XLM requires up to 200K updates to converge.

\begin{table*}[t]
\begin{center}
\vspace{-0.5em}
\caption{BLEU scores on the \emph{large scale} WMT'14 English-French (En-Fr), WMT'16 English-German (En-De) and WMT'16 English-Romanian (En-Ro) unsupervised translation tasks.}
\vspace{-0.5em}
\begin{tabular}{lcccccc}
\toprule
{\bf Method / Data}      & {\bf En-Fr} & {\bf Fr-En} & {\bf En-De} & {\bf De-En} & {\bf En-Ro}  & {\bf Ro-En}\\
\midrule
NMT \citep{lample2018phrase_unsup}                  & 25.1  & 24.2  & 17.2  & 21.0  & 21.1 & 19.4\\
PBSMT \citep{lample2018phrase_unsup}                & 27.8  & 27.2  & 17.7  & 22.6  & 21.3 & 23.0\\
Multi-agent dual learning \citep{multiagent}                   & --- & --- & 19.3 & 23.8 & --- & --- \\
XLM \citep{conneau2019cross_xlm}         & 33.4  & 33.3  & 26.4  & 34.3  & 33.3 & 31.8\\
MASS \citep{song2019mass}   & 37.5	& 34.9	& 28.3	& 35.2	& 35.2	& 33.1\\
\midrule
CBD                      & \textbf{38.2}    & \textbf{35.5}  & \textbf{30.1}  & \textbf{36.3}  & \textbf{36.3} & \textbf{33.8} \\
\bottomrule
\end{tabular}
\vspace{-1em}
\label{table:large_results}
\end{center}
\end{table*}

\subsection{Base WMT experiments}\label{subsec:wmt}

\paragraph{Setup.} Since the results in \cref{subsec:wmt_large} may have been influenced by large scale data and pretrained models, we then seek to evaluate the effectiveness of our CBD method in scenarios where none of the above conveniences are provided. 
Specifically, we use the News Crawl 2007-2008 datasets for {English (En)}, French (Fr) and German (De), and News Crawl 2015 dataset for Romanian (Ro), and limit the total number of sentences per language to 5M. This is, in fact, the default data setup in the code provided by \citet{lample2018phrase_unsup,conneau2019cross_xlm}.
For the NMT models, we follow \citet{lample2018phrase_unsup} to train the UMT agents with a parameter-shared Transformer \citep{vaswani2017attention} that has 6 layers and 512 dimensions and a batch size of 32 sentences. We use {joint} Byte-Pair Encoding (BPE) \citep{sennrich2015neural} and train fastText \citep{fasttext} on the BPE tokens to initialize the token embeddings. For the PBSMT \citep{pbsmt_Koehn:2003:SPT} models, {following \citet{lample2018phrase_unsup},} we use MUSE \citep{muse_conneau2017word} to generate the initial phrase table and run 4 iterations of back-translation. We subsample 500K sentences from the 5M monolingual sentences at each iteration to train the PBSMT models.\footnote{For PBSMT, \citet{lample2018phrase_unsup} subsampled 5M out of 193M sentences of monolingual data.\label{foot1}}. For XLM \citep{conneau2019cross_xlm}, we follow the same setup as described in \cref{subsec:wmt_large}, except that we pretrain and finetune the XLM model from scratch on the base 5M dataset. We choose the best model based on validation loss and use beam size of 5. We use a 4-GPU system to train the models. To ensure randomness in the PBSMT agents, we use different seeds for MUSE training and randomly subsample different sets of data during training. To achieve the same for neural agents (NMT and XLM), we simply use different seeds to initialize the models and sample batches of training data.

\begin{table}[h!]   
\begin{center}
\caption{BLEU scores on the \emph{base} WMT'14 English-French (En-Fr), WMT'16 English-German (En-De) and WMT'16 English-Romanian (En-Ro) unsupervised translation tasks.
}
\setlength{\tabcolsep}{3pt}
\begin{tabular}{lcccccc}
\toprule
{\bf Method}          & {\bf En-Fr} & {\bf Fr-En} & {\bf En-De} & {\bf De-En} & {\bf En-Ro}  & {\bf Ro-En}\\
\midrule
\textbf{Data}               & 5M  & 5M  & 5M & 5M & 3M  & 3M\\
\midrule
NMT                  & 24.7	& 24.5  & 14.5	& 18.2	& 16.7	& 16.3   \\
\hspace{0.5em} + CBD                                   & 26.6	& 25.7  & 16.6	& 20.5  & 18.1  & 17.8\\
\midrule
PBSMT                 & 17.1	& 16.4	& 10.9	& 13.6	& 10.5	& 11.7	\\
\hspace{0.5em} + CBD                          & 21.6	& 20.6	& 15.0	& 17.7  & 11.3	& 14.5  \\
\midrule
XLM                         & 33.0	& 31.5  & 23.9  & 29.3  & 30.6  & 27.9 \\
\hspace{0.5em} + CBD        & {35.4}	& {33.0}    & {26.1}  & {31.5}  & 32.2   & 29.2 \\
\bottomrule
\end{tabular}
\vspace{-1em}
\label{table:main_results2}
\end{center}
\end{table}

\paragraph{Results.} \Cref{table:main_results2} shows the experimental results of different UMT approaches with and without CBD. First of all, with the datasets {that are} 30-50 times smaller than the ones used in \cref{subsec:wmt_large}, the baselines perform around 2 to 3 BLEU worse than the scores reported in \citet{lample2018phrase_unsup,conneau2019cross_xlm} (see \Cref{table:large_results}). As shown, the CBD-enhanced model with the pretrained XLM achieves 35.4 and 33.0 BLEU on the WMT'14 En-Fr and Fr-En tasks respectively, which are 2.4 and 1.5 BLEU improvements over the baseline. {It also surpasses \citet{conneau2019cross_xlm} by 2.0 BLEU in En-Fr task}, despite the fact that their model was trained with 274M combined bilingual sentences (compared to our setup of {10M} sentences). CBD also boosts the scores for XLM in En-De, De-En, En-Ro, Ro-En by around 2.0 BLEU. For the NMT systems, CBD also outperforms the baselines by 1 to 2 BLEU. More interestingly, PBSMT models are known to be deterministic, but CBD is still able to improve data diversity and provide performance boost by up to 4.0 BLEU points.

\vspace{-0.5em}
\subsection{IWSLT experiments}\label{subsec:iwslt}
We also demonstrate the effectiveness of CBD on relatively small datasets for IWSLT En-Fr and En-De translation tasks. The IWSLT'13 En-Fr dataset contains 200K sentences for each language. We use the IWSLT15.TED.tst2012 set for validation and the IWSLT15.TED.tst2013 set for testing. The IWSLT'14 En-De dataset contains 160K sentences for each language. We split it into 95\% for training and 5\% for validation, and we use IWSLT14.TED.\{dev2010, dev2012, tst2010,tst1011, tst2012\} for testing. 
For these experiments, we use the neural UMT method \citep{lample2018phrase_unsup} with a Transformer of 5 layers and 512 model dimensions, and trained using only 1 GPU. 

From the results in \Cref{table:ablation_iwslt}, we can see that CBD improves the performance in all the four tasks by 2-3 BLEU compared to the NMT baseline of \cite{lample2018phrase_unsup}.



\begin{table}[h!]
\begin{center}
\caption{BLEU scores on the unsupervised IWSLT'13 English-French (En-Fr) and IWSLT'14 English-German (En-De) tasks.}
\vspace{-0.5em}
\begin{tabular}{lcccc}
\toprule
{\bf Method}          & {\bf En-Fr} & {\bf Fr-En} & {\bf En-De} & {\bf De-En} \\
\midrule
NMT                         & 29.6	& 30.7  & 15.8  & 19.1    \\
\hspace{1em} + CBD          & 31.8 & 31.8  & 18.4  & 21.7    \\
\bottomrule
\end{tabular}
\vspace{-1em}
\label{table:ablation_iwslt}
\end{center}
\end{table}





\vspace{-0.5em}
\section{Understanding CBD}\label{subsec:why_it_works}




\subsection{Cross-model back-translation is key} \label{subsec:cross_translation}

As mentioned, crucial to our strategy's success is the cross-model back-translation, where the agent operating at the first level must be different from the one in the second level. To verify this, we compare CBD with similar variants that do not employ the cross-model element in the WMT tasks. We refer to these variants {commonly} as \emph{back-translation distillation} (BD). The first variant \emph{BD (1,1)} has only 1 UMT agent that translates the monolingual data only once and uses these synthetic parallel pairs to distill the model $\theta$. The second variant \emph{BD (1,2)} employs 2 UMT agents, similar to CBD, to produce 2 sets of synthetic parallel data from the monolingual data and uses both of them for distillation. Finally, the third variant \emph{BD (2,2)} uses 2 UMT agents to sample translations from the monolingual data in forward and backward directions using the same respective agents. In other words, BD (2,2) follows similar procedures in \Cref{alg:macd}, except that it optimizes the following loss with $\theta_{\alpha}$ being alternated between $\theta_1$ and $\theta_2$:
\begin{equation}
\overline{\mathcal{L}_\theta}(\theta_{\alpha}) =\hspace{-0.7em} \displaystyle \mathop{\mathbb{E}}_{\substack{z_s \sim P(\cdot|y_t,\theta_{\alpha}), y_t \sim P(\cdot|x_s,\theta_{\alpha}), x_s \sim \sX_s\\z_t \sim P(\cdot|y_s,\theta_{\alpha}), y_s \sim P(\cdot|x_t, \theta_{\alpha}), x_t \sim \sX_t}} [\gJ(\theta)] \label{eqn:bd_loss}
\end{equation}
From the comparison in \Cref{table:similar_variants}, we see that none of the BD variants noticeably improves the performance across the language pairs, while CBD provides consistent gains of 1.0-2.0 BLEU. In particular, the BD (1,1) variant fails to improve as the distilled model is trained on the same synthetic data that the UMT agent is already trained on. 
The variant BD (1,2) is in fact similar in sprit to \citep{nguyen2019data}, which improves supervised and semi-supervised MT. However, it fails to do so in the unsupervised setup, {due to the lack of supervised agents.}
The variant BD (2,2) also fails because the 2{\text{nd}} level synthetic data is already optimized during iterative back-translation training of the UMT agents, leaving the distilled model with no extra information to exploit. Meanwhile, cross-model back-translation enables CBD to translate the second-level data by an agent other than the first-level agent. In this strategy, the second agent produces targets that the first agent is not aware of, while the second agent receives as input the sources that are foreign to it. This process creates corrupted but new information, which the supervised MT model can leverage to improve the overall MT performance through more data diversity.

\begin{table}[h!]   
\begin{center}
\caption{BLEU comparison of CBD vs. no cross-model variants in the \emph{base} WMT'14 English-French (En-Fr), WMT'16 English-German (En-De) and English-Romanian (En-Ro) tasks.
}
\setlength{\tabcolsep}{3pt}
\begin{tabular}{lcccccc}
\toprule
{\bf Method}          & {\bf En-Fr} & {\bf Fr-En} & {\bf En-De} & {\bf De-En} & {\bf En-Ro}  & {\bf Ro-En}\\
\midrule
NMT   & 24.7	& 24.5  & 14.5	& 18.2	& 16.7	& 16.3    \\
BD(1/1) & 24.5	& 24.5  & 14.0  & 17.5  & 16.1  & 15.9 \\
BD(1/2) & 24.6	& 24.6  & 14.1  & 17.8  & 16.4  & 16.2\\
BD(2/2) & 24.8	& 24.7  & 14.4  & 18.1  & 16.9  & 16.4\\
CBD     & \textbf{26.6}	& \textbf{25.7}  & \textbf{16.6}	& \textbf{20.5}  & \textbf{18.1}  & \textbf{17.8} \\
\bottomrule
\end{tabular}
\vspace{-1em}
\label{table:similar_variants}
\end{center}
\end{table}

\subsection{CBD produces diverse data} \label{subsec:data_diversity}

Having argued that cross-model back-translation creates extra information for the supervised MT model to leverage on, we hypothesize that such extra information can be measurable by the diversity of the generated data. To measure this, we compute the \emph{reconstruction BLEU} and compare the scores for BD (2,2) and CBD in the WMT En-Fr, En-De and En-Ro tasks.
The scores are obtained by using the first agent to translate the available monolingual data in language $s$ to $t$ and then the second agent to translate those translations back to language $s$. After that, a BLEU score is measured by comparing the reconstructed text with the original text. In BD, the first and second agents are identical, while they are distinct for CBD. From the results in \Cref{table:diversity_bleu}, we observe that the reconstruction BLEU scores of CBD are more than 10 points lower than those of BD, indicating that the newly generated data by CBD are more diverse and different from the original data.


\begin{table}[h!]
\begin{center}
\caption{Reconstruction BLEU scores of BD and CBD in different languages for the \emph{base} WMT unsupervised translation tasks. Lower BLEU means more diverse.}
\vspace{-0.5em}
\resizebox{\columnwidth}{!}{%
\setlength{\tabcolsep}{3pt}
\begin{tabular}{lcccccc}
\toprule
{\bf Method }          & {\bf En-Fr} & {\bf Fr-En} & {\bf En-De} & {\bf De-En} & {\bf En-Ro}  & {\bf Ro-En}\\
\midrule
BD                   & 76.0	& 72.4  & 75.3	& 63.7  & 73.2  & 71.5      \\
CBD                  & 63.1	& 59.7  & 60.3	& 50.5  & 61.1	& 56.9      \\
\bottomrule
\end{tabular}
}
\vspace{-1em}
\label{table:diversity_bleu}
\end{center}
\end{table}


In \Cref{table:diversity_duplicates}, we further report the ratio of duplicate source-target pairs to the amount of synthetic parallel data created by CBD. We sample 30M synthetic parallel data using the CBD strategy and examine the amount of duplicate pairs for the WMT En-Fr, En-De and En-Ro tasks.
We can notice that across the language pairs, only around 14\% of the parallel data are duplicates. 
Given that only about 5M (3.5M for En-Ro) sentences are \emph{real} sentences and the remaining 25M sentences are synthetic, this amount of duplicates is surprisingly low. This fact also explains why CBD is able to exploit extra information better than any standard UMT to improve the performance.

\begin{table}[h!]
\begin{center}
\caption{Comparison between the amount of real data, generated data by CBD and the duplicates per language pair for the \emph{base} WMT'14 En-Fr, WMT'16 En-De and En-Ro unsupervised MT tasks.}
\resizebox{\columnwidth}{!}{%
\begin{tabular}{lccc}
\toprule
{\bf Method }          & {\bf En-Fr} & {\bf En-De} & {\bf En-Ro} \\
\midrule
Real data               & 5M    & 5M    & 3.5M  \\
Generated data          & 30M   & 30M   & 29M   \\
Duplicate pairs         & 4.4M (14.5\%)   & 3.8M(12.7\%)     & 3.9M (13.4\%) \\
\bottomrule
\end{tabular}
}
\vspace{-1em}
\label{table:diversity_duplicates}
\end{center}
\end{table}


\subsection{Comparison with ensembles of models {and ensemble knowledge distillation}} \label{subsec:alternatives1}

Since CBD utilizes {outputs from} two UMT agents for supervised distillation, it is interesting to see how  it performs compared to an ensemble of UMT models and ensemble knowledge distillation \citep{ensemble_distill_freitag2017}.
To perform ensembling, we average the probabilities of the two UMT agents at each decoding step. {For ensemble distillation, we generate synthetic parallel data from an ensemble of UMT agents, which is then used to train the supervised model.}

\begin{table}[h!]
\begin{center}
\caption{BLEU comparison of CBD vs. an ensemble of UMT agents and ensemble knowledge distillation \citep{ensemble_distill_freitag2017} on \emph{base} WMT'14 En-Fr, WMT'16 En-De and En-Ro translation tasks.}
\resizebox{\columnwidth}{!}{%
\setlength{\tabcolsep}{3pt}
\begin{tabular}{lcccccc}
\toprule
{\bf Method}          & {\bf En-Fr} & {\bf Fr-En} & {\bf En-De} & {\bf De-En} & {\bf En-Ro}  & {\bf Ro-En}\\
\midrule
NMT Baseline                & 24.7	& 24.5  & 14.5	& 18.2	& 16.7	& 16.3    \\
Ensemble of 2 agents        & 25.2	& 24.8	& 15.3	& 19.1	& 17.7	& 17.1 \\
Ensemble distillation       & 17.3	& 20.0 	& 3.5	& 3.7	& 1.2	& 1.1 \\
CBD                         & \textbf{26.6}	& \textbf{25.7}  & \textbf{16.6}	& \textbf{20.5}  & \textbf{18.1}  & \textbf{17.8} \\
\bottomrule
\end{tabular}
}
\vspace{-1em}
\label{table:compare_with_ensemble}
\end{center}
\end{table}

From the results on the WMT translation tasks in \Cref{table:compare_with_ensemble}, we observe that ensembles of models improve the performance only by 0.5-1.0 BLEU, while CBD provides larger gains (1.0-2.0 BLEU) across all the tasks. 
These results demonstrate that CBD is capable of leveraging the potentials of multiple UMT agents better than how an ensemble of agents does.
This is in contrast to {data diversification} \citep{nguyen2019data}, which is shown to mimic and perform similarly to model ensembling. More importantly, during inference, an ensemble of models requires more memory and computations (twice in this case) to store and execute multiple models. In contrast, CBD can throw away the UMT teacher agents after training and needs only one single model for inference. Meanwhile, ensemble knowledge distillation \citep{ensemble_distill_freitag2017}, which performs well with supervised agents, performs poorly in unsupervised MT tasks. The reason could be that the UMT agents may not be suitable for the method {originally} intended for supervised learning. 
{Further inspection in the Appendix suggests that many samples in the ensemble translations contain incomprehensible repetitions.}

\begin{table}[t!]
\begin{center}
\caption{Comparison with other alternatives on the \emph{base} WMT En-Fr, Fr-En, En-De and De-En, with XLM as the base model.}
\setlength{\tabcolsep}{3pt}
\begin{tabular}{lcccc}
\toprule
{\bf WMT}              & {\bf En-Fr}    & {\bf Fr-En}  & {\bf En-De} & {\bf De-En} \\
\midrule
XLM                              & 33.0	& 31.5  & 23.9  & 29.3  \\
\midrule
Sampling (temp=0.3)                  & 33.5 & 32.2  & 24.3	& 30.2 \\
Top-k sampling                              & 33.18 & 32.26 & 24.0 & 29.9\\
Top-p sampling                            & \multicolumn{4}{c}{Diverge}\\
Target noising                              & 32.8 & 30.7  & 24.0	& 29.6 \\
Multi-agent dual learning                   & 33.5 & 31.7  & 24.6   & 29.9\\
\midrule
CBD                                  & {35.4}	& {33.0}    & {26.1}  & {31.5}  \\
\bottomrule
\end{tabular}
\vspace{-1em}
\label{table:compare_with_others}
\end{center}
\end{table}

\subsection{Comparison with other potential alternatives} \label{subsec:alternatives2}

In this section, we compare CBD with other alternatives in the  text generation literature that also attempt to increase diversity. {While many of these methods are generic, we adopt them in the UMT framework and compare their performance with our CBD technique in the WMT En-Fr, Fr-en, En-De, and De-en tasks, taking the XLM \citep{conneau2019cross_xlm} as the base model.} 

One major group of alternatives is \emph{sampling} based methods, where the model samples translations following multinomial distributions during iterative back-translation. Specifically, we compare the CBD with \Ni sampling with temperature 0.3 \citep{understanding_backtranslation_scale,hier_neural_story_gen}, \Nii top-k sampling \citep{radford2019language_gpt2}, and \Niii nucleus or top-p sampling \citep{curious_text_degeneration}. Plus, we compare CBD with \emph{target noising}, where we add random noises to the translations of the UMT model during iterative back-translation. Finally, multi-agent dual learning \citep{multiagent} is also considered as another alternative, where multiple unsupervised agents are used train the end {supervised} model.

The results are reported in \Cref{table:compare_with_others}. We can see that while the sampling based methods indeed increase the diversity significantly, they do not improve the performance as much as CBD does. The reason could be that the extra data generated by {(stochastic)} sampling are noisy and their quality is not as good as deterministic predictions from the two UMT agents via cross-model back-translation. On the other hand, target noising does not provide a consistent improvement while multi-agent dual learning  achieves less impressive gains compared to CBD.

\subsection{Translationese effect}\label{sec:translationese}

It can be seen that our cross-model back-translation method is indeed a modified version of back-translation \citep{backtranslate_sennrich-etal-2016-improving}. Therefore, it is necessary to test if this method suffers from the \emph{translationese effect} \citep{eval_back_translation_translationese}. As pointed out in their work, back-translation only shows performance gains with translationese source sentences but does not improve when the sentences are natural text.\footnote{Translationese is human translation of a natural text by a professional translator. Translationese tends to be simpler, more grammatically correct, but lacks contextual sentiments and fidelity.} \citet{nguyen2019data} show that the translationese effect only exhibits in a semi-supervised setup, where there are both parallel and monolingual data. However, while they show that their supervised back-translation technique is not impacted by the translationese effect, they left out the question whether unsupervised counterparts are affected.

Therefore, we test our unsupervised CBD method against the translationese effect by conducting the same experiment. More precisely, we compare the BLEU scores of our method versus the XLM baseline \citep{conneau2019cross_xlm} in the WMT'14 English-German test sets in the three setups devised by \citet{eval_back_translation_translationese}: 

\vspace{-1em}
\begin{itemize}[itemsep=-0.3pt]
    \item Natural source $\rightarrow$ translationese target ($X\rightarrow Y^*$).
    \item Translationese source $\rightarrow$ natural target ($X^*\rightarrow Y$)
    \item Translationese of translationese of source to translationese of target ($X^{**}\rightarrow Y^*$).
\end{itemize}
\vspace{-1em}

\Cref{table:translationese} shows that our method outperforms the baseline significantly in the natural source $\rightarrow$ translationese target scenario ($X\rightarrow Y^*$), while it may not improve the translationese source scenario ($X^*\rightarrow Y$) considerably. The results demonstrate that our method behaves differently than what the translationese effect indicates. More importantly, the translations of the natural source sentences are improved, which indicates the practical usefulness of our method. Furthermore, in line with the findings in \citet{nguyen2019data}, the experiment shows that the translationese effect may only exhibit in a semi-supervised setup, but not in supervised or unsupervised setups.

\begin{table}[t!]
\begin{center}
\caption{BLEU scores of CBD and the baseline \citep{conneau2019cross_xlm} on the translationese effect \citep{eval_back_translation_translationese}, in the \emph{base} WMT'14 English-German setup.}
\setlength{\tabcolsep}{3pt}
\begin{tabular}{lccc}
\toprule
{\bf WMT'14 En-De} & {\bf $X\rightarrow Y^*$}     & {\bf $X^*\rightarrow Y$}      & {\bf $X^{**}\rightarrow Y^*$} \\
\midrule
XLM Baseline           & 18.63    & 18.01   & 25.59  \\
CBD      & 20.40  &  18.31  &  27.72\\
\bottomrule
\end{tabular}
\vspace{-1em}
\label{table:translationese}
\end{center}
\end{table}

\vspace{-0.5em}
\section{Related work} \label{sec:relwork}


The first step towards utilizing the vast monolingual data to boost MT quality is through semi-supervised training.
Back-translation \citep{backtranslate_sennrich-etal-2016-improving,understanding_backtranslation_scale} is an effective approach to exploit target-side monolingual data. Dual learning \citep{duallearning_he2016dual,multiagent}, meanwhile, trains {backward and forward models} concurrently and intertwines them together. Recently, \citet{mirror_Zheng2020Mirror-Generative} proposed a variational method to couple the translation and language models {through a shared latent space}.
{There have also been attempts in solving low-resource translation problems with limited parallel data \citep{uni_nmt_low_resource_gu2018universal,hallucinating_phrase_lowrs_mt_irvine2014hallucinating,flores}.} In the realm of SMT, cross-lingual dictionaries have been used to reduce parallel data reliance \citep{end2end_smt_small_data_irvine2016end,smt_without_parallel_klementiev2012toward}.

In recent years, unsupervised word-translation via cross-lingual word embedding has seen a huge success  \citep{muse_conneau2017word,bilingual_word_embed_artetxe2017learning,Artetxe-2018-acl}.
This opened the door for UMT methods that employ the three principles described in \cref{sec:background}. \citet{lample2017unsupervised} and \citet{unmt_artetxe2018unsupervised} were among the first of this kind, who use denoising autodecoder for language modeling and iterative back-translation.
\citet{lample2017unsupervised} use MUSE \citep{muse_conneau2017word} as the initialization to bootstrap the model, while \citet{unmt_artetxe2018unsupervised} use cross-lingual word embeddings \citep{bilingual_word_embed_artetxe2017learning}.
\citet{lample2018phrase_unsup} later suggested the use of BPE \citep{sennrich2015neural} and fastText \citep{fasttext} to initialize the model and the parameters sharing. Pretrained language models \citep{devlin2018bert} are then used to initialize the entire network \citep{conneau2019cross_xlm}. \citet{song2019mass} proposed to pretrain an encoder-decoder model while \citet{unsup_smt_artetxe-etal-2018-unsupervised} suggested a combination of PBSMT and NMT with subword information.\footnote{Since \citet{unsup_smt_artetxe-etal-2018-unsupervised} did not provide the code, we were unable to apply CBD to their work.} Plus, pretraining {BART \citep{lewis-etal-2020-bart}} on multi-lingual corpora improves the initialization process \citep{mbart_liu2020multilingual}.


Our proposed CBD works outside the three-principle UMT framework and is considered as an add-on to any underlying UMT system. There exist some relevant approaches to CBD. First, it is similar to \citet{nguyen2019data}, which generates a diverse set of data from multiple \emph{supervised} MT agents. Despite being effective in supervised and semi-supervised settings, a direct implementation of it in UMT underperforms due to lack of supervised signals (\cref{subsec:cross_translation}). In order to successfully exploit unsupervised agents, CBD requires cross-model back-translation which is the key to its effectiveness. 

Second, CBD can be viewed as an augmentation technique \citep{data_aug_low_resource_fadaee-etal-2017-data,switchout}. Although the denoising autoencoding built in a typical UMT system also performs augmentation, the noising process is rather naive, while CBD augments data by well-trained agents. Sampling based methods are also considered data diversification strategies, where the model samples translation tokens not by greedy selection (taking $\argmax$ of probabilities), but by a predefined multinomial distribution. Simple sampling with temperature is often used in many text generation tasks \citep{understanding_backtranslation_scale,hier_neural_story_gen}. More advanced top-k sampling \citep{radford2019language_gpt2} is used in GPT-2, where a subset of the vocabulary is selected and re-scaled to compute probabilities. Meanwhile, top-p sampling \citep{curious_text_degeneration} is used to tackled text degeneration. Our CBD method draws a clear distinction from these methods in that the presumed extra synthetic data is generated not by a random stochastic process, but by well-trained models through the cross-translation procedure.

Third, CBD is related to ensembling \citep{when_net_disaggree_perrone1992networks} and ensemble knowledge distillation \citep{knowledge_distill_kim_rush_2016,ensemble_distill_freitag2017}. Ensembling \citep{when_net_disaggree_perrone1992networks} refers to a type of inference strategies, where multiple differently trained models are used to predict the output probabilities given an input, which are then averaged out to acquire the final output. Ensemble knowledge distillation \citep{knowledge_distill_kim_rush_2016,ensemble_distill_freitag2017}, meanwhile, use multiple models to perform ensemble inference to generate one-way synthetic targets from the original source data, which are then used to distill the final model. The major difference between our method and the aforementioned ensembling methods is that they seek to produce the most accurate translations with less variance, while ours seeks to produce as much diverse data as possible. Along with the fact that these distillation schemes are currently applied to supervised settings only, the results in \Cref{table:compare_with_ensemble} may indicate that they are not suitable for unsupervised machine translation.
Similar to our method, multi-agent dual learning \citep{multiagent} also uses multiple models in both forward and backward directions, but the data is generated in an ensembling style and its objective to minimize the reconstruction losses instead of to generate diverse synthetic data.


\vspace{-0.5em}
\section{Conclusion}

We have proposed cross-model back-translated distillation (CBD) - a method that works outside the three existing principles for unsupervised MT and is applicable to any UMT methods. CBD establishes the state of the art in the unsupervised WMT'14 English-French, WMT'16 English-German and English-Romanian translation tasks.
It also outperforms the baselines in the IWSLT'14 German-English and IWSLT'13 English-French tasks by up to 3.0 BLEU. Our analysis shows that CBD embraces data diversity and extracts more model-specific intrinsic information than what an ensemble of models would do.


\section*{Acknowledgements}
We deeply appreciate the efforts of our anonymous reviewers and meta-reviewer in examining and giving us feedback on our paper. We also thank Prathyusha Jwalapuram for proofreading the paper. Xuan-Phi Nguyen is supported by the A*STAR Computing and Information Science (ACIS) scholarship, provided by the Agency for Science, Technology and Research Singapore (A*STAR). Shafiq Joty would like to thank the funding support from NRF (NRF2016IDM-TRANS001-062), Singapore.





\bibliography{main}
\bibliographystyle{icml2021}

\section{Appendix}


In the following supplementary material, we first provide the full mathematical derivations of the loss function $\mathcal{L}$ presented in the paper (\cref{sec:appendix:derivations}). Then, we provide the generalized version of our method cross-model back-translated distillation, or GCBD, and measure its effectiveness in the IWSLT English-German, German-English, English-French and French-English unsupervised tasks (\cref{sec:appendix:generalized_version}). In addition, we investigate why ensemble knowledge distillation \citep{ensemble_distill_freitag2017}, which boosts the performance in a supervised setup, fails to do so in an unsupervised setup where we replace the supervised agents used in the method with the UMT agents (\cref{sec:appendix:analysis_duplicates}). 
{Finally, in \cref{sec:appendix:compare_sup_mt}, we provide a comparison between unsupervised models and supervised counterparts to provide a perspective of how far unsupervised machine translation research has progressed.}

\subsection{Derivations of negative log likelihood $\mathcal{N}(\theta_{\alpha},\theta_{\beta})$}\label{sec:appendix:derivations}

\begin{figure}
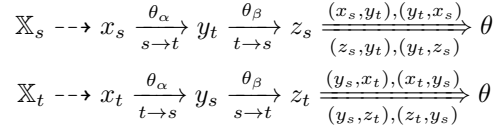

\begin{center}
\begin{equation*}
    \sX_s \dashrightarrow x_s  \xrightarrow[s \rightarrow t]{\theta_{\alpha}}  y_t \xrightarrow[t \rightarrow s]{\theta_{\beta}} z_s \xRightarrow[(z_s,y_t),(y_t,z_s)]{(x_s,y_t),(y_t,x_s)} \theta
\end{equation*}
\vspace{-1em}
\begin{equation*}
    \sX_t \dashrightarrow x_t  \xrightarrow[t \rightarrow s]{\theta_{\alpha}}  y_s \xrightarrow[s \rightarrow t]{\theta_{\beta}} z_t \xRightarrow[(y_s,z_t),(z_t,y_s)]{(y_s,x_t),(x_t,y_s)} \theta
\end{equation*}
\vspace{-1em}
\caption{The sampling process of $x_s,x_t,y_s,y_t,z_s,z_t$.}
\label{fig:supp:cbd:translation_diagram} 
\end{center}
\end{figure}

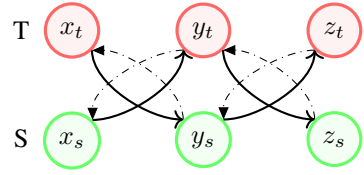
\begin{figure}
\vspace{-0.5em}
    \centering
    \begin{tikzpicture}[
        roundnodeg/.style={circle, draw=green!60, fill=green!5, very thick, minimum size=7mm},
        squarenoder/.style={rectangle, draw=red!60, fill=red!5, very thick, minimum size=7mm},
        roundnoder/.style={circle, draw=red!60, fill=red!5, very thick, minimum size=7mm},
        squarenodeg/.style={rectangle, draw=green!60, fill=red!5, very thick, minimum size=7mm},
        ]
        \node                   (target)                                {T};
        \node                   (source)    [below=2.7em of target]     {S};
        \node[roundnoder]       (targetx)   [right=0.2em of target]      {$x_t$};
        \node[roundnoder]       (targety)   [right=of targetx]      {$y_t$};
        \node[roundnoder]       (targetz)   [right=of targety]      {$z_t$};
        \node[roundnodeg]       (sourcex)   [below=2em of targetx]       {$x_s$};
        \node[roundnodeg]       (sourcey)   [below=2em of targety]      {$y_s$};
        \node[roundnodeg]       (sourcez)   [below=2em of targetz]      {$z_s$};
        %
        \draw[->][thick] (sourcex.north east) .. controls +(right:4.5mm) and +(down:4.5mm) .. (targety.south west);
        \draw[->][thick] (targety.south east) .. controls +(down:4.5mm) and +(left:4.5mm) .. (sourcez.north west);
        \draw[->][thick] (targetx.south east) .. controls +(down:4.5mm) and +(left:4.5mm) .. (sourcey.north west);
        \draw[->][thick] (sourcey.north east) .. controls +(right:4.5mm) and +(down:4.5mm) .. (targetz.south west);
        \draw[arrows={-Triangle[angle=45:5pt]}][dash dot] (targety.south west) .. controls +(left:4.5mm) and +(up:4.5mm) .. (sourcex.north east);
        \draw[arrows={-Triangle[angle=45:5pt]}][dash dot] (sourcey.north west) .. controls +(up:4.5mm) and +(right:4.5mm) .. (targetx.south east);
        \draw[arrows={-Triangle[angle=45:5pt]}][dash dot] (targetz.south west) .. controls +(left:4.5mm) and +(up:4.5mm) .. (sourcey.north east);
        \draw[arrows={-Triangle[angle=45:5pt]}][dash dot] (sourcez.north west) .. controls +(up:4.5mm) and +(right:4.5mm) .. (targety.south east);
    \end{tikzpicture}
    \caption{Graphical model representation of CBD. The model ($\theta$) is trained on all parallel pairs (shown as directed links): $(x_s, y_t), (y_t, x_s),$ $(y_t, z_s), (z_s, y_t), (x_t, y_s), (y_s, x_t), (y_s, z_t), (z_t, y_s)$.}
    \label{fig:supp:graphical:gen}
\end{figure}

In this section, we provide the complete mathematical derivations of the loss function $\mathcal{L}$ in the paper. Recalling that we are supposed to maximize the log probabilities of the variables $x_s$, $y_t$, $z_s$, $x_t$, $y_s$ and $z_t$ according to the sampling process in \Cref{fig:supp:cbd:translation_diagram} and the graphical model in \Cref{fig:supp:graphical:gen}. Otherwise speaking, we seek to minimize the following negative log likelihood:
\begin{equation}
    \hspace{-0.2em}{\gJ}(\theta) = -\log P_{\theta}(x_s,y_t,z_s) - \log P_{\theta}(x_t,y_s,z_t) \hspace{-1em}\label{eqn:supp:nll}
\end{equation}
Then we can expand the first term as follows:
\begin{equation}
\begin{split}
\hspace{-1.2em}&\log P_{\theta}(x_s,y_t,z_s) = \log \frac{P_{\theta}(x_s,y_t,z_s)}{P_{\theta}(x_s,y_t)}P_{\theta}(x_s,y_t)\\
\hspace{-1.2em}&   \hspace{0.5em} = \log P_{\theta}(z_s|x_s,y_t) + \log P_{\theta}(x_s,y_t) \hspace{-1em} \\
\hspace{-1.2em}&   \hspace{0.5em} = \log P_{\theta}(z_s|x_s,y_t) + \log \frac{P_{\theta}(x_s,y_t)}{P_{\theta}(y_t)}P_{\theta}(y_t) \hspace{-1em} \\
\hspace{-1.2em}&   \hspace{0.5em} = \log P_{\theta}(z_s|x_s,y_t) + \log P_{\theta}(x_s|y_t) +  \log P_{\theta}(y_t) \hspace{-1em}
\end{split}
\label{eqn:supp:der:1:mid}
\end{equation}
Since $z_s$ is independent from $x_s$ given $y_t$ according to the graphical model (\cref{fig:supp:cbd:translation_diagram}), we have $P_{\theta}(z_s|x_s,y_t)=P_{\theta}(z_s|y_t)$, then Eq. \ref{eqn:supp:der:1:mid} can be reduced to:
\begin{equation}
\begin{split}
&\log P_{\theta}(x_s,y_t,z_s) = \log P_{\theta}(z_s|y_t) + \log P_{\theta}(x_s|y_t)\\
&\hspace{9em}   + \log P_{\theta}(y_t)
\end{split}
\label{eqn:supp:der:1}
\end{equation}
Alternatively, the first term can also be express as follows:
\begin{equation}
\begin{split}
\hspace{-1.2em}&\log P_{\theta}(x_s,y_t,z_s) = \log P_{\theta}(z_s|x_s,y_t) + \log P_{\theta}(x_s,y_t) \hspace{-1em} \\
\hspace{-1.2em}&   \hspace{1em} = \log P_{\theta}(z_s|y_t) + \log P_{\theta}(y_t,x_s) \hspace{-1em} \\
\hspace{-1.2em}&   \hspace{1em} = \log \frac{P_{\theta}(y_t|z_s)P_{\theta}(z_s)}{P_{\theta}(y_t)} + \log \frac{P_{\theta}(y_t,x_s)}{P_{\theta}(x_s)} P_{\theta}(x_s) \hspace{-1em}\\
\hspace{-1.2em}&   \hspace{1em} = \log P_{\theta}(y_t|z_s) + \log P_{\theta}(z_s) - \log P_{\theta}(y_t) \hspace{-1em}\\
\hspace{-1.2em}&   \hspace{1.5em}  + \log P_{\theta}(y_t|x_s) + \log P_{\theta}(x_s) \hspace{-1em}
\end{split}
\label{eqn:supp:der:2}
\end{equation}
After that, we expand the second term in similar fashion, which we yield:
\begin{equation}
\begin{split}
&\log P_{\theta}(x_t,y_s,z_t) = \log P_{\theta}(z_t|y_s) + \log P_{\theta}(x_t|y_s)\\
&\hspace{9em}   + \log P_{\theta}(y_s) 
\end{split}
\label{eqn:supp:der:3}
\end{equation}
\vspace{-1em}
\begin{align}
&\log P_{\theta}(x_t,y_s,z_t) = \log P_{\theta}(y_s|z_t) + \log P_{\theta}(y_s|x_t) \nonumber\\
&\hspace{3em} + \log P_{\theta}(z_t) + \log P_{\theta}(x_t) - \log P_{\theta}(y_s) \label{eqn:supp:der:4}
\end{align}
Then, by adding up Eq. \ref{eqn:supp:der:1}, \ref{eqn:supp:der:2}, \ref{eqn:supp:der:3} and \ref{eqn:supp:der:4} together, and then divide it by 2, we will derive the negative log likelihood of Eq. \ref{eqn:supp:nll} as:
\begin{equation}
\begin{split}
\hspace{-0.5em}&{\gJ}(\theta) =\frac{1}{2} [-\log P_{\theta}(y_t|z_s) - \log P_{\theta}(y_t|x_s) \hspace{-0.5em}\\
\hspace{-0.5em}& - \log P_{\theta}(z_s|y_t) - \log P_{\theta}(x_s|y_t) - \log P_{\theta}(y_s|z_t)  \hspace{-0.5em}\\
\hspace{-0.5em}& - \log P_{\theta}(y_s|x_t) - \log P_{\theta}(z_t|y_s) - \log P_{\theta}(y_s|x_t)  \hspace{-0.5em}\\
\hspace{-0.5em}& - \log P_{\theta}(x_s) - \log P_{\theta}(z_s) - \log P_{\theta}(x_t) - \log P_{\theta}(z_t) ]\hspace{-0.5em}
\end{split}
\hspace{-0.5em}
\label{eqn:supp:nll:out}
\end{equation}
Considering the generation process of $x_s,y_s,z_s,x_t,y_t$ and $z_t$, we minimize the following CBD loss function $\mathcal{L}$:
\begin{equation}
\hspace{-0.3em}\mathcal{L}_{\theta}(\theta_{\alpha},\theta_{\beta}) =\hspace{-0.7em} \displaystyle \mathop{\mathbb{E}}_{\substack{z_s \sim P(\cdot|y_t,\theta_{\beta}), y_t \sim P(\cdot|x_s,\theta_{\alpha}), x_s \sim \sX_s\\z_t \sim P(\cdot|y_s,\theta_{\beta}), y_s \sim P(\cdot|x_t, \theta_{\alpha}), x_t \sim \sX_t}}[{\gJ}(\theta)]\hspace{-0.7em} \label{eqn:cbd_loss}
\end{equation}
where $\theta_{\alpha},\theta_{\beta} \in \Theta$ are input parameters, which are specified in the CBD training procedure described in the main paper. Note $\theta_{\alpha}$ is used to sample $y_t,y_s$ from $x_s,x_t$ while $\theta_{\beta}$ is used to back-translate $y_t,y_s$ to $z_s,z_t$ respectively.

It is note-worthy that in practice, we do not explicitly optimize the non-conditional terms $P_{\theta}(x_s)$, $P_{\theta}(x_t)$, $P_{\theta}(z_s)$ and $P_{\theta}(z_t)$. The reason is that the MT model $\theta$ is built as a strictly cross-lingual model, which means that it can only translate from one language to another, and possibly vice versa. It is not, how, equipped to train an explicit language model that only aims to optimize non-conditional log probabilities. We did attempt to pseudo-optimize them by using denoising-autoencoding strategy in the preliminary experiments. The results, however, yield no difference and sometimes under-performance. We conjecture that this is due to technical difficulty in forcing a single-language model loss upon a cross-lingual model for the sole purpose of improving machine translation. We put this in our future work.

\subsection{Generalized version}\label{sec:appendix:generalized_version}

In this section, we describe a generalized version of our CBD, which involves multiple UMT agents instead of just two. Then, we test this method in the IWSLT experiments to demonstrate its effectiveness and characteristics. Specifically, in addition to the input monolingual data $\sX_s$ and $\sX_t$ of languages $s$ and $t$ and the supervised model $\theta$, we introduce another hyper-parameter $n$ to indicate the number of unsupervised agents used to perform cross-model back-translation. The generalized cross-model back-translated distillation (GCBD) strategy is presented in \Cref{alg:gcbd}. In this method, instead of training only two agents, the method trains a set of $n$ UMT agents $\Theta=\{\theta_1,...,\theta_n\}$. During training, we iteratively select two orderly distinct agents $\theta_i$ and $\theta_j$ from $\Theta$ and use them to perform cross-model back-translation and train the model $\theta$.

\begin{algorithm}[tb]
\footnotesize
\caption{Generalized Cross-model Back-translated Distillation (GCBD): Given monolingual data $\sX_{s}$ and $\sX_{t}$ of languages $s$ and $t$, and hyper-parameter $n$, return a UMT model with parameters $\theta$.}
\label{alg:gcbd}
\begin{algorithmic}[1]
    \FOR{$i \in 1,...,n$}
        \STATE Train UMT agent with parameters $\theta_i$
    \ENDFOR
    \STATE Initialize MT model $\theta$ (randomly or with pretrained model)
    \WHILE{until convergence}
        \FOR{$i \in 1,...,n$}
            \FOR{$j \in 1,...,n$ where $j \neq i$}
                \STATE $\theta \gets \theta - \eta \nabla_{\theta} \mathcal{L}(\theta_{\alpha}=\theta_i,\theta_{\beta}=\theta_j)$
            \ENDFOR
        \ENDFOR
    \ENDWHILE
    \STATE \textbf{return} $\theta$
\end{algorithmic}
\end{algorithm}

To evaluate GCBD in comparison with CBD, we conduct experiments with the IWSLT'13 English-French (En-Fr) and IWSLT'14 English-German (En-De) tasks. The setup for these experiments are identical to the IWSLT experiment in the main paper, except that we vary the hyper-parameter $n=(2,3,4)$ to determine the optimal number of agents. The results are reported in \Cref{table:gcbd_iwslt}. As it can be seen, increasing the number of agents $n$ to 3 adds an additional $0.4-1.0$ BLEU improvement compared to the standard CBD. Moreover, using 4 UMT agents does not improve the performance over using just 3 UMT, despite that this setup still outperforms the standard CBD. The results indicate that increasing the system complexity further is not always optimal and diminishing return is observed as we add more agents to the system.

\begin{table}[t!]
\begin{center}
\caption{BLEU scores on the unsupervised IWSLT'13 English-French (En-Fr) and IWSLT'14 English-German (En-De) tasks with varying number of agents $n$ of GCBD.}
\setlength{\tabcolsep}{2pt}
\begin{tabular}{lcccc}
\toprule
{\bf Method}          & {\bf En-Fr} & {\bf Fr-En} & {\bf En-De} & {\bf De-En} \\
\midrule
NMT    & 29.6	& 30.7  & 15.8  & 19.1    \\
\hspace{0.5em} + GCBD ($n=2$) (CBD)            & 31.8 & 31.8  & 18.4  & 21.7    \\
\hspace{0.5em} + GCBD ($n=3$)                  & \textbf{32.8} & \textbf{32.1}  & \textbf{19.2}	& \textbf{22.2}       \\
\hspace{0.5em} + GCBD ($n=4$)                  & 32.3 & 32.0	& 19.1	& 21.9 \\
\bottomrule
\end{tabular}
\vspace{-1em}
\label{table:gcbd_iwslt}
\end{center}
\end{table}


\subsection{Analysis of degeneration in ensemble knowledge distillation}\label{sec:appendix:analysis_duplicates}

Ensemble knowledge distillation \citep{ensemble_distill_freitag2017} has been used to enhance supervised machine translation. It uses multiple strong (supervised) teachers to generate synthetic parallel data from both sides of the parallel corpora by averaging the decoding probabilities of the teachers at each step. The synthetic data are then used to train the student model. 
Having seen its effectiveness in the supervised setup, we apply this same tactic to unsupervised MT tasks by replacing the supervised teachers with unsupervised MT agents. However, the method surprisingly causes drastic performance drop in the WMT'14 En-Fr, WMT'16 En-De and En-Ro unsupervised MT tasks.

By manual inspection, we found that many instances of the synthetic data are incomprehensible and contain repetitions, {which is a degeneration behavior}. We then quantitatively measure the percentage of sentences in the synthetic data containing tri-gram repetitions by counting the number of sentences where a word/sub-word is repeated at least three consecutive times. 
As reported in the main paper, from 30\% to 86\% of the synthetic data generated {by} the ensemble knowledge distillation (Ens-Distil) method are incomprehensible and contain repetitions. 
{Relative to the performance of CBD, the performance drop in ensemble distillation is also more dramatic for language pairs with higher percentage of degeneration (En-Ro and En-De).} This explains why the downstream student model fails to learn from these corrupted data. The results indicate that UMT agents are unable to jointly translate through ensembling strategy the monolingual data  that they were trained on. This phenomenon may require further research to be fully understood. 
On the other hand, with less than 0.1\% tri-gram repetitions, CBD generates little to no repetitions, which partly explains why it is able to improve the performance.


\begin{table}[t!]
\caption{Percentage of tri-gram repetitions in the synthetic data generated by ensemble knowledge distillation \citep{ensemble_distill_freitag2017}, compared to those created by CBD; and the respective test BLEU scores in the \emph{base} WMT'14 En-Fr, WMT'16 En-De and En-Ro unsupervised tasks.}
\setlength{\tabcolsep}{2pt}
\begin{tabular}{lcccccc}
\toprule
{\bf Method}          & {\bf En-Fr} & {\bf Fr-En} & {\bf En-De} & {\bf De-En} & {\bf En-Ro}  & {\bf Ro-En}\\
\midrule
\multicolumn{7}{l}{\bf \% tri-gram repetition}\\
Ens-Distil  & $30.3$\%	& $34$\% 	& $73$\%	& $76$\%	& $43$\%	& $86$\% \\
CBD               & $10^{-3}$\%	& $10^{-2}$\%	& $10^{-2}$\%	& $10^{-1}$\%	& $10^{-2}$\%	& $10^{-2}$\%	\\
\midrule
\multicolumn{7}{l}{\bf BLEU on test set}\\
Ens-Distil  & 17.3	& 20.0 	& 3.5	& 3.7	& 1.2	& 1.1 \\
CBD               & 26.6	& 25.7  & 16.6	& 20.5  & 18.1  & 17.8\\

\bottomrule
\end{tabular}%
\label{table:gcbd_iwslt}%
\end{table}

\subsection{Convergence curves of CBD compared with the baselines}

\begin{figure}[t!]
    \centering
    \includegraphics[width=\columnwidth]{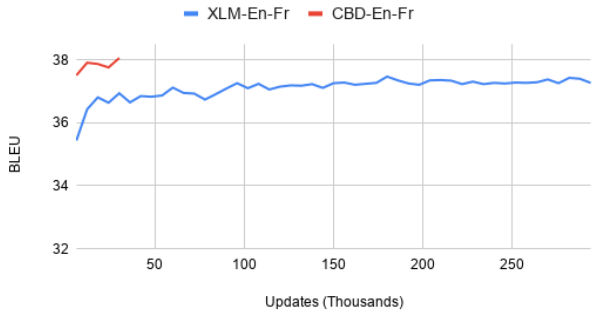}
    \caption{Convergence speed of CBD in comparison with baseline XLM, represented by the test BLEU score of the \emph{WMT En-Fr} task after a given number of training updates.}
    \label{fig:convergence_enfr}
    \vspace{-1em}
\end{figure}

\begin{figure}[t!]
    \centering
    \includegraphics[width=\columnwidth]{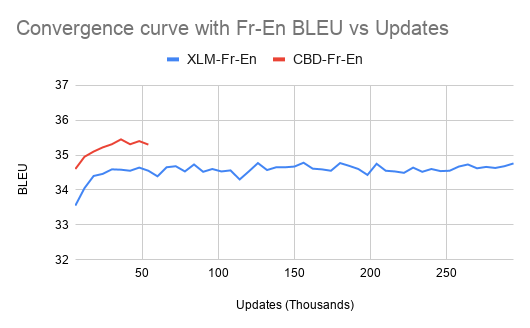}
    \caption{Convergence speed of CBD in comparison with baseline XLM, represented by the test BLEU score of the \emph{WMT Fr-En} task after a given number of training updates.}
    \label{fig:convergence_fren}
    \vspace{-1em}
\end{figure}

\begin{figure}[t!]
    \centering
    \includegraphics[width=\columnwidth]{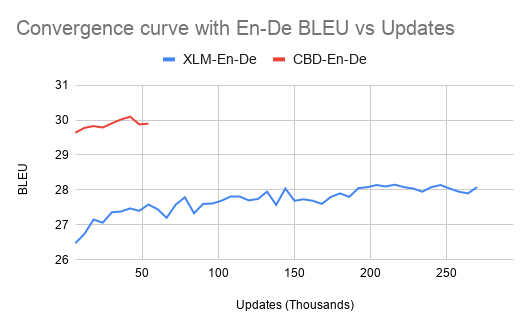}
    \caption{Convergence speed of CBD in comparison with baseline XLM, represented by the test BLEU score of the \emph{WMT En-De} task after a given number of training updates.}
    \label{fig:convergence_ende}
    \vspace{-1em}
\end{figure}

\begin{figure}[t!]
    \centering
    \includegraphics[width=\columnwidth]{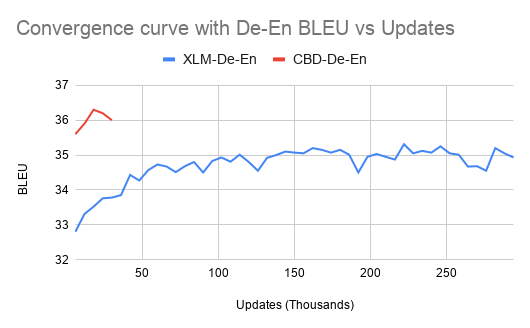}
    \caption{Convergence speed of CBD in comparison with baseline XLM, represented by the test BLEU score of the \emph{WMT De-en} task after a given number of training updates.}
    \label{fig:convergence_deen}
    \vspace{-1em}
\end{figure}

\begin{figure}[t!]
    \centering
    \includegraphics[width=\columnwidth]{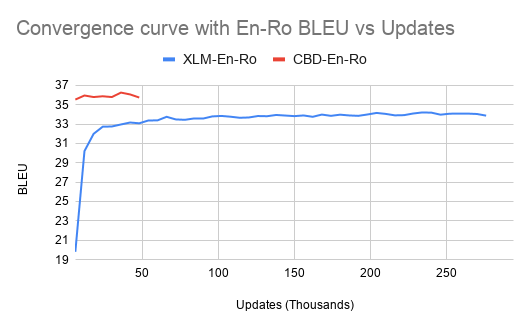}
    \caption{Convergence speed of CBD in comparison with baseline MASS, represented by the test BLEU score of the \emph{WMT En-Ro} task after a given number of training updates.}
    \label{fig:convergence_enro}
    \vspace{-1em}
\end{figure}

\begin{figure}[t!]
    \centering
    \includegraphics[width=\columnwidth]{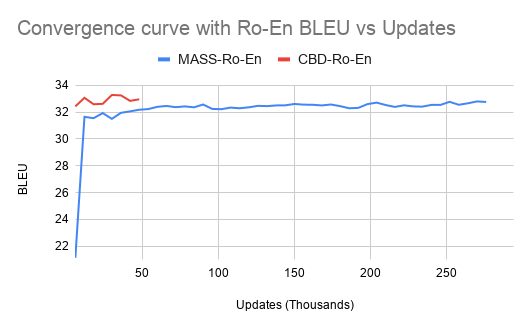}
    \caption{Convergence speed of CBD in comparison with baseline MASS, represented by the test BLEU score of the \emph{WMT Ro-En} task after a given number of training updates.}
    \label{fig:convergence_roen}
    \vspace{-1em}
\end{figure}

This section provides extra convergence curve charts for all 6 of the language pairs in the large scale WMT English-French (\Cref{fig:convergence_enfr} \& \Cref{fig:convergence_fren}), WMT English-German (\Cref{fig:convergence_ende} \& \Cref{fig:convergence_deen}) and WMT English-Romanian (\Cref{fig:convergence_enro} \& \Cref{fig:convergence_roen}) tasks. As it can be seen from the charts, CBD converges rapidly and outperforms the baselines with little additional resources, given the pretrained models provided by \citet{conneau2019cross_xlm} and \citet{song2019mass}.


\subsection{Comparison with supervised MT}\label{sec:appendix:compare_sup_mt}

\begin{table}[h!]
\begin{center}
\caption{BLEU scores on the WMT'14 English-French (En-Fr) and WMT'16 English-German (En-De) tasks of unsupervised MT methods (MASS and CBD), in comparison to supervised MT method \citep{scaling_nmt_ott2018scaling}.}
\vspace{-0.5em}
\begin{tabular}{lcc}
\toprule
{\bf Method}          & {\bf En-Fr} & {\bf En-De} \\
\midrule
\multicolumn{3}{l}{Unsupervised MT} \\
\midrule
XLM \citep{conneau2019cross_xlm}    & 33.4   & 26.4  \\
MASS \citep{song2019mass}   & 37.5	& 28.3 \\
CBD                      & 38.2     & 30.1 \\
\midrule
\multicolumn{3}{l}{Supervised MT} \\
\midrule
Transformer \citep{scaling_nmt_ott2018scaling}  & 43.2  & 33.0 \\
\bottomrule
\end{tabular}
\vspace{-1em}
\label{table:compare_sup}
\end{center}
\end{table}

In this section, we compare the performances of the CBD method, along with previous SOTA unsupervised models, with the standard supervised Transformer model \citep{scaling_nmt_ott2018scaling} to present a perspective of how much progress the field of unsupervised machine translation has made. More specifically, we use the provided Transformer models \emph{pretrained} on the parallel WMT'14 English-French and English-German datasets and evaluate them on the WMT'14 En-Fr and WMT'16 En-De test sets, as similarly done for unsupervised counterparts. The results are presented in \Cref{table:compare_sup}. As it can be seen, unsupervised MT models have made significant advancement throughout multiple iterations and refinement \citep{conneau2019cross_xlm,song2019mass}. However, while the CBD method further improve the performance, it still lags behind the supervised MT model \citep{scaling_nmt_ott2018scaling} by around 3 to 5 BLEU points.

\end{document}